\documentclass[10pt,twocolumn,twoside]{IEEEtran}
\usepackage{cite}
\usepackage{amsmath}
\usepackage{algorithmic}
\usepackage{array}
\usepackage{url}
\usepackage{multirow}
\usepackage{subfigure}
\usepackage{tabularx}
\usepackage{verbatim}
\usepackage{graphicx}
\usepackage{amssymb}
\usepackage{caption2}
\usepackage{balance} 
\usepackage{color}
\begin{document}
%
\title{Fine-grained Discriminative Localization via WSDL}
\title{Multiple Fine-grained Discriminative Localization via Saliency-guided Multi-scale Faster R-CNN}
\title{Fast Fine-grained Image Classification via \\ Weakly Supervised Discriminative Localization}

\author{Xiangteng He, Yuxin Peng and Junjie Zhao
\thanks{This work was supported by National Natural Science Foundation of China
under Grants 61771025 and 61532005.}
\thanks{The authors are with the Institute of Computer Science and Technology,
Peking University, Beijing 100871, China. Corresponding author: Yuxin Peng (e-mail: pengyuxin@pku.edu.cn).}}


\maketitle

\begin{abstract}
Fine-grained image classification is to recognize hundreds of subcategories in each basic-level category. Existing methods employ discriminative localization to find the key distinctions among similar subcategories.
However, existing methods generally have two limitations: 
(1) Discriminative localization relies on region proposal methods to hypothesize the locations of discriminative regions, which are \emph{time-consuming} and the \emph{bottleneck} of classification speed.
(2) The training of discriminative localization depends on object or part annotations, which are heavily \emph{labor-consuming} and the \emph{obstacle} of marching towards practical application.
It is highly challenging to address the two key limitations \emph{simultaneously}, and existing methods only focus on one of them.
Therefore, we propose a \emph{weakly supervised discriminative localization approach (WSDL) for fast fine-grained image classification} to address the two limitations at the same time, and its main advantages are: 
(1) \emph{$n$-pathway end-to-end discriminative localization network} is designed to improve classification speed, which simultaneously localizes multiple different discriminative regions for one image to boost classification accuracy, and shares full-image convolutional features generated by region proposal network to accelerate the process of generating region proposals as well as reduce the computation of convolutional operation.
(2) \emph{Multi-level attention guided localization learning} is proposed to localize discriminative regions with different focuses automatically, without using object and part annotations, avoiding the labor consumption. Different level attentions focus on different characteristics of the image, which are complementary and boost the classification accuracy.
Both are jointly employed to simultaneously improve classification speed and eliminate dependence on object and part annotations.  
Compared with state-of-the-art methods on 2 widely-used fine-grained image classification datasets, our WSDL approach achieves both the best accuracy and efficiency of classification.
\end{abstract}


%
\IEEEpeerreviewmaketitle

\section{Introduction}

\begin{figure}[!t]
    \begin{center}\includegraphics[width=1\linewidth]{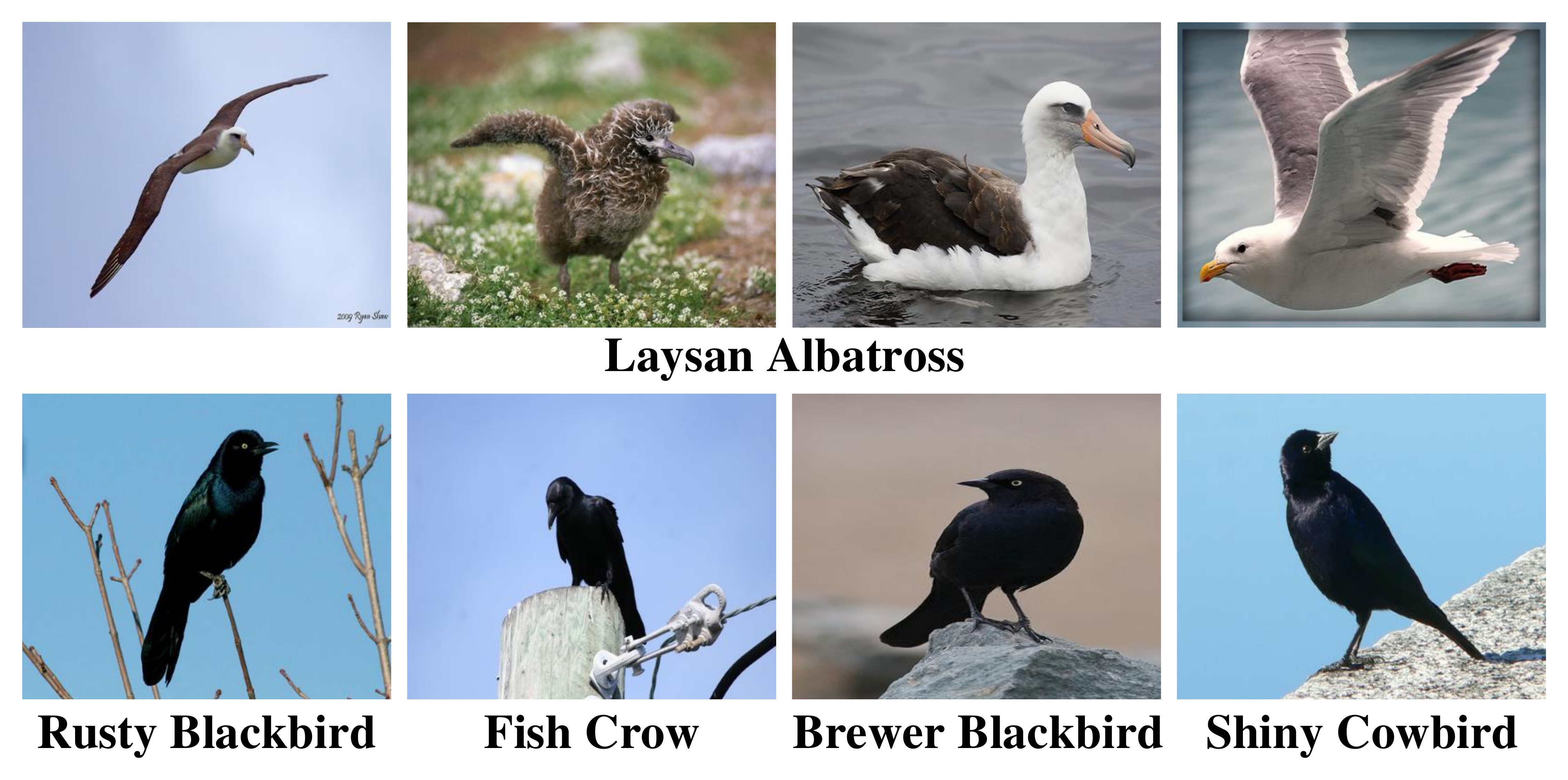}
    \caption{Examples of CUB-200-2011 dataset \cite{wah2011caltech}. Large variance in the same subcategory is shown in the first row, and small variance among different subcategories is shown in the second row.}
    \label{interintra}
    \end{center}
\end{figure}

\IEEEPARstart{F}{ine-grained} image classification aims to recognize hundreds of subcategories in the same basic-level category, which lies in the continuum between basic-level image classification (e.g. object recognition \cite{lowe1999object,Girshick_2015_ICCV}) and identification of individuals (e.g. face recognition \cite{sudha2011self,paier2017hybrid}).
It is one of the most significant and highly challenging open problems of computer vision due to the following two aspects:
(1) \emph{Large variance in the same subcategory.} As shown in the first row of Fig. \ref{interintra}, the four images belong to the same subcategory of ``Laysan Albatross'', but they are different in poses, views, feathers and so on. It is easy for human beings to classify them into different subcategories.
(2) \emph{Small variance among different subcategories.} As shown in the second row of Fig. \ref{interintra}, the four images belong to different subcategories, but they are all black and look similar. It is hard for human beings to distinguish ``Fish Crow'' from the other three subcategories.
These subcategories in the same basic-level category look similar in global appearance, but distinct in some discriminative regions of the objects, such as the head. So the localization of the key discriminative regions becomes crucial for fine-grained image classification. Recently, methods of fine-grained image classification based on discriminative localization have achieved great progress \cite{zhang2014part,zhang2014fused,krause2015fine,xiao2015application,zhang2016picking,simon2015neural,he2017spatial}.

Fine-grained image classification has wide applications in automatic driving, biological conservation, cancer detection and so on.
In the process of converting technology into application, there are two important problems that need to be solved urgently:
(1) \emph{Time consumption.} Some existing methods mainly focus on achieving better classification accuracy, but ignore the problem of time consumption. However, real-time performance is one of the most important criteria in the applications of fine-grained image classification, which satisfies the response speed requirements of users.
(2) \emph{Labor consumption.}
The annotations of image (e.g. the image-level subcategory label, the bounding box of the object and part locations) are required in the training phase of many existing methods, and even in the testing phase. While the annotations are labor-consuming and unrealistic in the applications of fine-grained image classification. So utilizing as few annotations as possible is the key point to convert fine-grained image classification into applications.

Early works only focus on achieving better classification accuracy, but ignore aforementioned two problems.
Zhang et al. \cite{zhang2014part} propose a Part-based R-CNN method for fine-grained image classification, which learns whole-object and part detectors with geometric constraints between them.
The learning phase of detectors depends on the annotations of image-level subcategory label, object and part. They first generate thousands of region proposals for every image via Selective Search method \cite{uijlings2013selective}, which is one of the most popular region proposal methods, and greedily merges pixels based on engineered low-level features. Then they utilize the learned whole-object and part detectors to detect object and parts from the generated region proposals, and finally predict a fine-grained subcategory from a pose-normalized representation. This framework is widely used in fine-grained image classification.
Krause et al. \cite{krause2015fine} adopt the box constraint of Part-based R-CNN \cite{zhang2014part} to train part detectors using only object annotations. These methods rely on region proposal methods implemented with CPU to hypothesize the locations of discriminative regions, which are \emph{time-consuming} and the \emph{bottleneck} of classification speed. And the training of their discriminative localization depends on object or part annotations, which are heavily \emph{labor-consuming} and the \emph{obstacle} of marching towards practical application. 
It is highly challenging and significant to address the two problems \emph{simultaneously}, and the existing methods only focus on one of them, while ignoring the other one.

\emph{For addressing the problem of time consumption}, researchers focus on designing end-to-end network and avoiding the application of the time-consuming region proposal methods implemented with CPU.
Zhang et al. propose a Part-stacked CNN architecture \cite{zhang2016spda}, which consists of a fully convolutional network and a two-stream classification network. They first utilize the fully convolutional network to localize discriminative regions, and then adopt the two-stream classification network to encode object-level and part-level features simultaneously. Part-stacked CNN is over two orders of magnitude faster than Part-based R-CNN \cite{zhang2014part}, but requires the annotations of image-level subcategory label, object and part in the training phase, which are \emph{labor-consuming}. 

\emph{For addressing the problem of labor consumption}, researchers focus on the localization of the discriminative regions under the weakly supervised setting, which denotes that neither object nor part annotations are used in both training and testing phases. 
Xiao et al. propose a two-level attention model \cite{xiao2015application}: object-level attention is to select region proposals relevant to a certain object, and part-level attention is to localize discriminative parts. 
It is the first work to classify fine-grained images without using object or part annotations in both training and testing phases, but still achieve promising results \cite{zhang2016weakly}. 
Simon and Rodner propose a constellation model \cite{simon2015neural} to localize discriminative regions of object, leveraging CNN to find the constellations of neural activation patterns. A part model is estimated by selecting part detectors via constellation model. And then the part model is used to extract features for classification. 
These methods do not depend on object or part annotations, but they generally utilize Selective Search \cite{uijlings2013selective} method to generate region proposals, which is \emph{time-consuming}. 

Different from them, this paper proposes a \emph{weakly supervised discriminative localization approach (WSDL) for fast fine-grained image classification}, which is the first attempt based on discriminative localization to simultaneously improve fine-grained image classification speed and eliminate dependence on object and part annotations. Its main contributions can be summarized as follows:
\begin{itemize}
\item
\emph{\textbf{$n$-pathway end-to-end discriminative localization network}} is designed to improve classification speed, which consists of multiple localization networks and one region proposal network. The localization networks simultaneously localize different discriminative regions for one image to boost classification accuracy, and share full-image convolutional features generated by region proposal network to reduce the computation of convolutional operation so that classification speed is improved. 
\item
\emph{\textbf{Multi-level attention guided localization learning}} is proposed to localize discriminative regions with different focuses automatically, without using object and part annotations, avoiding the labor consumption. Different level attentions focus on different characteristics of the image, carrying multi-grained and multi-level information. They are complementary with each other, and their combination further boosts the classification accuracy.
\end{itemize}

Our previous conference paper \cite{hemm2017} proposes a discriminative localization approach via saliency-guided Faster R-CNN, which localizes the discriminative region in the image to boost the classification accuracy.
The main differences between the proposed WSDL approach and our previous conference paper \cite{hemm2017} can be summarized as the following two aspects:
(1) Our previous conference paper \cite{hemm2017} applies one level attention, while our WSDL approach further employs multi-level attention to guide the discriminative localization learning, which localizes multi-grained and multi-level discriminative regions to boost fine-grained classification. Their combination boosts the classification accuracy due to the complementarities among them.
(2) Our WSDL approach designs $n$-pathway network structure to reduce the growth of time consumption in classification. Time consumption is reduced by sharing full-image convolutional features among different localization networks with different level attentions.
The architecture in \cite{hemm2017} can only deal with one level attention, and the application of multi-level attention will cause the linear growth of time consumption in classification.  
Compared with state-of-the-art methods on 2 widely-used fine-grained image classification datasets, the effectiveness of our WSDL approach is verified, where our WSDL approach achieves both the best accuracy and efficiency of classification.

The rest of this paper is organized as follows: 
Section \uppercase\expandafter{\romannumeral2} briefly reviews the related works on fine-grained image classification and object detection, 
Section \uppercase\expandafter{\romannumeral3} presents our WSDL approach in detail, and Section \uppercase\expandafter{\romannumeral4} introduces the experimental results as well as the experimental analyses. Finally Section \uppercase\expandafter{\romannumeral5} concludes this paper and presents the future works of this paper.

\section{Related Work}
In this section, we review the related works on fine-grained image classification and object detection.
\subsection{Fine-grained Image Classification}
Fine-grained image classification is one of the most fundamental and challenging open problems in computer vision, and has drawn extensive attention in both academia and industry. Early works \cite{xie2014spatial,gao2014learning} focus on the design of feature representation and classifier based on the basic low-level descriptors, such as SIFT \cite{lowe2004distinctive}. However, the performance of these methods is limited due to the handcrafted features. Recently, deep learning has achieved great success in the domains of computer vision, speech recognition, natural language processing and so on. Inspired by this, many researchers begin to study on the problem of fine-grained image classification by deep learning \cite{xiao2015application,he2017spatial,zhang2014part,zhang2016weakly,zhang2016picking}, and have achieved great progress. 

Since the discriminative characteristics generally localize in the regions of the object and parts, most existing works generally follow the two-stage pipeline: First localize the object and parts, and then extract their features to train classifiers. For the first stage, some works \cite{xie2013hierarchical,berg2013poof} directly utilize the human annotations (i.e. the bounding box of the object and part locations) to localize the object and parts.
Since the human annotations are labor-consuming, some researchers begin to only utilize them in the training phase. Zhang et al. propose the Part-based R-CNN \cite{zhang2014part} to directly utilize the object and part annotations to learn the whole-object and part detectors with geometric constraints between them. This framework is widely used in fine-grained image classification. 

Recently, fine-grained image classification methods begin to focus on how to achieve promising performance without using any object or part annotations. The first work under such weakly supervised setting is the two-level attention model \cite{xiao2015application}, which utilizes the attention mechanism of the CNNs to select region proposals corresponding to the object and parts, and achieves promising results even compared with those methods relying on the object and part annotations. Inspired by this work, Zhang et al. \cite{zhang2016picking} incorporate deep convolutional filters for both parts selection and description. He and Peng \cite{he2017spatial} integrate two spatial constraints for improving the performance of parts selection. 

\begin{figure*}[!ht]
    \begin{center}\includegraphics[width=1\linewidth]{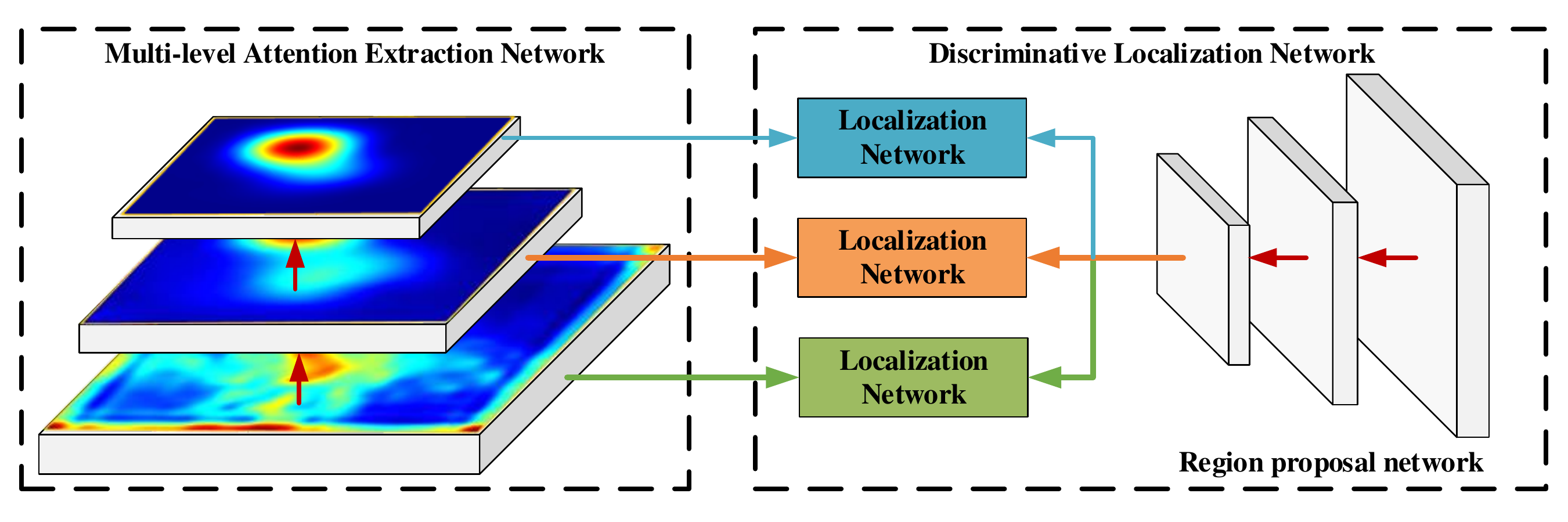}
    \caption{An overview of our WSDL approach. Multi-level attention extraction network (MAEN) extracts the attention information from multiple convolutional layers to provide the bounding boxes of discriminative regions for training discriminative localization network (DLN). The DLN consists of one region proposal network and multiple localization networks. We utilize 3-level attention and 3 localization networks in the figure to clearly demonstrate our WSDL approach.}
    \label{framework}
    \end{center}
\end{figure*}
\subsection{Object Detection}
Object detection is one of the most fundamental and challenging open problems in computer vision, which not only recognizes the objects but also localizes them in the images.
Like fine-grained image classification, early works are mainly based on basic low-level features, such as SIFT \cite{lowe2004distinctive} and HOG \cite{dalal2005histograms}. However, from 2010 onward, the progress of object detection based on these handcrafted features slows down. Due to the great success of deep learning in the competition of ImageNet LSVRC-2012, deep learning has been widely employed in computer vision, including object detection. We divide the object detection methods based on CNNs into 2 groups by the annotations used: (1) Supervised object detection, which needs the ground truth bounding box of the object. (2) Weakly supervised object detection, which does not need the ground truth bounding box of the object, and only needs image-level labels.

\subsubsection{Supervised Object Detection}
Girshick et al. \cite{girshick2014rich} propose a simple and scalable detection framework, regions with
CNN features, called R-CNN. First, it utilizes the region proposal method (i.e. Selective Search \cite{uijlings2013selective}) to generate thousands of region proposals for each image. 
Then it trains CNNs end-to-end to extract highly discriminative features of these region proposals.
Finally, it classifies these region proposals based on their discriminative features to determine whether the region proposal can be output as a bounding box of the object in the image. 
Inspired by R-CNN, many works follow the pipeline: First utilize the region proposal methods to generate region proposals for each image, and then employ CNNs to extract their features and classify their category.

However, these methods have a limitation: Time consumption is high because each region proposal needs to pass forward CNNs respectively, while each image generally generates thousands of region proposals. This limitation causes that object detection cannot satisfy the requirement of real-time performance.
For addressing this limitation, SPP-net \cite{he2015spatial} and Fast R-CNN \cite{Girshick_2015_ICCV} are proposed.
SPP-net applies a spatial pyramid pooling (SPP) layer to pool a fixed-length feature representation of each region proposal, which extracts the feature maps from the entire image only once, and avoids the time-consuming convolutional operation of each region proposal. Compared with R-CNN, SPP-net is 24 $\sim$ 102 $\times$ faster in object detection.
However, SPP-net cannot update the convolutional layers before the spatial pyramid pooling layer, and its extracted features need to be stored to disk, which limits both the accuracy and efficiency.
Therefore, Fast R-CNN \cite{Girshick_2015_ICCV} is proposed to fix the disadvantages of R-CNN and SPP-net. Fast R-CNN utilizes a region of interest (RoI) pooling layer to extract a fixed-length feature vector for each region proposal based on the feature map, employs multi-task loss to train the network in a single-stage, and updates all network layers in the training phase.
Compared with SPP-net, Fast R-CNN is 10 $\times$ faster and more accurate in object detection.

However, all the above methods are based on the region proposal methods, such as Selective Search \cite{uijlings2013selective}, EdgeBoxes \cite{zitnick2014edge}, which become the computational bottleneck. The region proposal methods are implemented with CPU, which causes that the time consumption of generating region proposals is high.
Therefore, Faster R-CNN \cite{ren2015faster} proposes a region proposal network (RPN) to generate region proposals and implements it with GPU, which makes the computation of generating region proposals nearly cost-free. 

\subsubsection{Weakly Supervised Object Detection}
All the above object detection methods need the ground truth bounding box of the object in the training phase, which is labor-consuming. Recently, many methods \cite{oquab2015object,oquab2014learning,zhou2016cvpr} begin to exploit weakly supervised object detection based on CNNs.
Oquab et al. \cite{oquab2014learning} design a method to reuse layers trained on the ImageNet 1K dataset \cite{imagenet_cvpr09} to compute mid-level representation, and localize the region of object based on the mid-level representation. 
Oquab et al. \cite{oquab2015object} further propose a weakly supervised learning method based on an end-to-end CNN only with image-level labels, and utilize max pooling operation to generate the feature map to localize the object.
Zhou et al. \cite{zhou2016cvpr} use global
average pooling (GAP) in CNN to generate the attention map for each image. Based on the attention map, the region of the object can be localized.


\section{Weakly Supervised Discriminative Localization}
We propose \emph{a weakly supervised discriminative localization approach (WSDL) for fast fine-grained image classification}, where an $n$-pathway end-to-end network is designed to localize discriminative regions and encode discriminative features simultaneously. Despite achieving a notable classification accuracy, our WSDL approach improves classification speed and eliminates dependence on object and part annotations simultaneously. An overview of our approach is shown as Fig. \ref{framework}.
It consists of two subnetworks: multi-level attention extraction network (MAEN) and discriminative localization network (DLN). 
Multi-level attention extraction network extracts multi-level attention information from different convolutional layers for each image, and generates multiple initialized discriminative regions based on the attention information. Then the bounding boxes of these discriminative regions are adopted as the annotations to guide the training of discriminative localization network, which localizes multiple discriminative regions that are significant for classification and avoids the dependence on object and part annotations. 
Both MAEN and DLN can generate the discriminative regions, but with different advantages:
(1) Instead of using the labor-consuming human annotations, MAEN provides the bounding box information of discriminative regions for the training of DLN automatically, even though the discriminative region is not very accurate. It is noted that MAEN is only employed in the training phase.
(2) Based on the initialized discriminative regions generated by MAEN, DLN further optimizes the learned discriminative regions to find where are more discriminative for distinguishing this subcategory from others. 
Their combination makes the best of their advantages and fixes their disadvantages to further achieve better classification performance.

\subsection{Multi-level Attention Extraction Network}
\begin{figure}[!t]
    \begin{center}\includegraphics[width=1\linewidth]{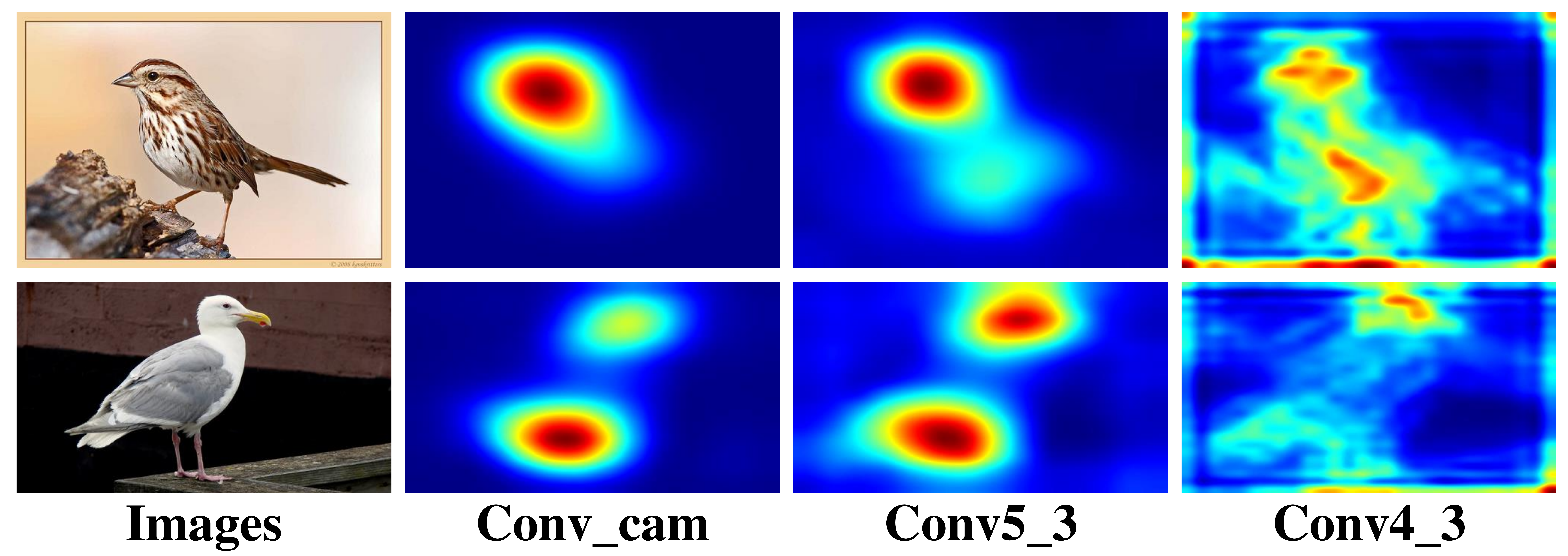}
    \caption{Examples of attention maps extracted by MAEN in our WSDL approach.}
    \label{saliencymap}
    \end{center}
\end{figure}

Attention is a behavioral and cognitive process of selectively concentrating on a discrete aspect of information \cite{anderson1990cognitive}. 
Tsotsos et al. state that visual attention mechanism seems to involve the selection of regions of interest in the visual field\cite{tsotsos1995modeling}.
And Karklin et al. indicate that neurons in primary visual cortex (e.g. V1) respond to the edge over a range of positions, and neurons in higher visual areas (e.g. V2 and V4) are more invariant to image properties and might encode shape \cite{karklin2009emergence}. The discovery is also shown in convolutional neural networks (CNNs), different feature maps (attention) reflect different characteristics of the image, as shown in Fig. \ref{saliencymap}. The images in different rows of Fig. \ref{saliencymap} are the attention maps extracted from the convolutional layers of ``Conv4\_3'', ``Conv5\_3'' and ``Conv\_cam'' in our multi-level attention extraction network respectively.
We can observe that different convolutional layers have different focuses, and provide complementary information to boost the classification accuracy.

According to the studies on visual attention mechanism, we design the multi-level attention extraction network (MAEN) to generate bounding box information for discriminative localization network. 
We take resized images as inputs and output $n$ feature maps from $n$ convolutional layers as the multi-level attention maps to indicate the importance of each pixel in the image for classification. Then we generate the bounding boxes of discriminative regions based on these attention maps. 
Inspired by CAM \cite{zhou2016cvpr},
we remove the fully-connected layers before the final output in CNN and replace them with global average pooling followed by a fully-connected softmax layer.
Then we sum the feature maps of the certain convolutional layer with weights to generate the attention map for each image. In this stage, we will generate $n$ attention maps based on $n$ different convolutional layers.
Finally, we perform binarization operation on each attention map with an adaptive threshold, which is obtained by OTSU algorithm \cite{otsu1979threshold}, and take the bounding box that covers the largest connected area as the discriminative region. Similarly, we obtain $n$ bounding boxes of discriminative regions, which are adopted as the bounding box information of discriminative localization network. 

For a given image $I$, we generate $n$ attention maps, the value of spatial location $(x,y)$ in $i$-th attention map is defined as follow:
\begin{gather}
M_i(x,y) = \sum \limits_{u_i} w_{u_i} f_{u_i}(x,y)
\end{gather}
where $M_i(x,y)$ indicates the importance of activation at spatial location $(x,y)$ for classification, $f_{u_i}(x,y)$ denotes the activation of neuron $u_i$ in the $i$-th convolutional layer at spatial location $(x,y)$, and $w_{u_i}$ denotes the weight used to sum the activation $f_{u_i}(x,y)$ to generate the attention map. For different convolutional layers, $w_{u_i}$ has different definitions as follow:
\begin{gather}
w_{u_i} = 
\begin{cases}
w_{u_i}^c, & last \ convolutional \ layer \\
\frac{1}{|u_i|}, & otherwise
\end{cases} 
\end{gather}
where $w_{u_i}^c$ denotes the weight corresponding to subcategory $c$ for neuron $u_i$ in the last convolutional layer, denoted as ``Conv\_cam'' in our MEAN.
We use the predicted result as the subcategory $c$ instead of using the image-level subcategory label. $|u_i|$ denotes the total number of neurons in $i$-th convolutional layer. 

\subsection{Discriminative Localization Network}
From multi-level attention extraction network, we obtain $n$ attention maps to guide the training of discriminative localization network. 
To make the best of the complementarity of multi-level attention information, we design an $n$-pathway end-to-end network based on Faster R-CNN \cite{ren2015faster}, which consists of multiple localization networks and one region proposal network.
Faster R-CNN is proposed to accelerate the process of detection as well as achieve promising detection performance.
We modify the original Faster R-CNN in two aspects: 
(1) The training phase of Faster R-CNN needs ground truth bounding box of discriminative region in the image, which is heavily labor-consuming. In this paper, we use the bounding box information provided by multi-level attention extraction network as the ground truth bounding box information, which eliminates the dependence on object and part annotations.
(2) Inspired by the discoveries on visual attention mechanism, we apply multi-level attention into our WSDL approach. 
However, the application of multi-level attention is restricted by the architecture of the original Faster R-CNN. Original Faster R-CNN consists of one region proposal network and one localization network, which restricts it to only localize one discriminative region at one time. We need to train $n$ Faster R-CNN models to apply the multi-level attention, which causes the linear growth of time consumption in classification. Therefore, we design an $n$-pathway end-to-end network with multiple localization networks and one region proposal network, where all the localization networks share the same full-image convolutional features generated by region proposal network.
 
Instead of using time-consuming region proposal methods such as Selective Search method \cite{uijlings2013selective}, region proposal network (RPN) is designed in Faster R-CNN to quickly generate region proposals for each image by sliding a small network over the feature map of last shared convolutional layer. At each sliding-window location, $k$ region proposals are simultaneously predicted, and they are parameterized relative to $k$ anchors. We apply $9$ anchors with $3$ scales and $3$ aspect ratios. For training RPN, a binary class label of being an object or not is assigned to each anchor, which depends on the Intersection-over-Union (IoU) \cite{everingham2015pascal} overlap with a ground truth bounding box of the object. But in our WSDL approach, we compute the IoU overlap with the bounding boxes of discriminative regions generated by MAEN rather than the ground truth bounding box of the object. 
And the loss function for an image is defined as:
\begin{gather}
L(\{p_i\},\{t_i\})=\frac{1}{N_{cls}} \sum_i L_{cls}(p_i,p_i^*)\nonumber \\ + \lambda \frac{1}{N_{reg}} \sum_i {p_i^*L_{reg}(t_i,t_i^*)}
\end{gather}
where $i$ denotes the index of an anchor in a mini-batch, $p_i$ denotes the predicted probability of anchor $i$ being a discriminative region, $p_i^*$ denotes the label of being a discriminative region of object or not depending on the bounding box $t_i^*$ generated by MAEN, $t_i$ is the predicted bounding box of discriminative region, $L_{cls}$ is the classification loss defined by log loss, and $L_{reg}$ is the regression loss defined by the robust loss function (smooth $L_1$). 

Since we apply multi-level attention into our WSDL approach, we employ $n$ localization networks, and each of them is the same with Fast R-CNN \cite{Girshick_2015_ICCV}.
All the localization networks are connected to RPN by a region of interest (RoI) pooling layer, which is employed to extract a fixed-length feature vector from feature map for each region proposal generated by RPN. Each feature vector is taken as the input of each localization network, and passes forward to generate two outputs: one is predicted subcategory and the other is predicted bounding box of discriminative region. 
Through discriminative localization network, we obtain the discriminative regions with multi-level attention. Then we average the predicted scores of the discriminative regions with multi-level attention and the original image to obtain the subcategory of each image.

\subsection{Training of MAEN and DLN}
MAEN learns the multi-level attention information of image to tell which regions are important and discriminative for classification, and then guides the training of DLN.
RPN in DLN generates region proposals relevant to the discriminative regions of images. Considering that training RPN needs bounding boxes of discriminative regions provided by MAEN, we adopt the strategy of sharing convolutional weights between MAEN and RPN to promote the localization learning.

First, we train the MAEN. This network is first pre-trained on the ImageNet 1K dataset \cite{imagenet_cvpr09}, and then fine-tuned on the fine-grained image classification dataset, such as CUB-200-2011 dataset \cite{wah2011caltech}. Second, we train the RPN. Its initial weights of convolutional layers are cloned from MAEN. Instead of fixing the shared convolutional layers, all layers are fine-tuned in the training phase. 
Third, we train the localization networks. Since all the localization networks share full-image convolutional features generated by RPN, we fix the parameters of RPN when training the localization networks with multi-level attention.

\section{Experiments}
We conduct experiments on 2 widely-used datasets in the fine-grained image classification task: CUB-200-2011 \cite{wah2011caltech} and Cars-196 \cite{krause20133d} datasets. Our WSDL approach is compared with state-of-the-art methods to verify its effectiveness, where our WSDL approach achieves both the best accuracy and efficiency of fine-grained image classification.

\subsection{Datasets}
Two datasets are adopted in the experiments:
\begin{itemize}
\item
\textbf{CUB-200-2011} \cite{wah2011caltech} is the most widely-used dataset in fine-grained image classification task, which contains 11788 images of 200 subcategories belonging to the category of bird, 5994 images in the training set and 5794 images in the testing set. Each image is labeled with detailed annotations including an image-level subcategory label, a bounding box of the object and 15 part locations. In our experiments, only image-level subcategory label is used in the training phase.
\item
\textbf{Cars-196} \cite{krause20133d} contains 16185 images of 196 car subcategories, which is divided as follows: the training set contains 8144 images, and the testing set contains 8041 images. For each subcategory, 24$\sim$84 images are selected for training and 24$\sim$83 images for testing. Every image is annotated with an image-level subcategory label and a bounding box of the object. The same with CUB-200-2011 dataset, only image-level subcategory label is used in the training phase. 
\end{itemize}

\subsection{Evaluation Metrics}
\textbf{Accuracy} is adopted to comprehensively evaluate the classification performances of our WSDL approach as well as the compared state-of-the-art methods, which is widely used in fine-grained image classification \cite{zhang2016weakly,zhang2016picking,zhang2014part}, and defined as follow:
\begin{gather}
Accuracy = \frac{R_a}{R}
\end{gather} 
where $R$ denotes the number of images in the testing set (e.g. $R$ equals to 5794 in the CUB-200-2011 dataset), and $R_a$ denotes the number of images that are correctly classified. 

\textbf{Intersection-over-Union (IoU)} \cite{everingham2015pascal} is adopted to evaluate whether the predicted bounding box of discriminative region is a correct localization, and its definition is as follow: 
\begin{gather}
IoU = \frac{area(B_p \cap B_{gt})}{area(B_p \cup B_{gt})}
\end{gather}
where $B_p$ denotes the predicted bounding box of discriminative region, $B_{gt}$ denotes the ground truth bounding box of the object, $B_P \cap B_{gt}$ denotes the intersection of the predicted
and ground truth bounding boxes, and $B_p \cup B_{gt}$ denotes their union. The predicted bounding box of discriminative region is correctly localized, if the $IoU$ exceeds 0.5.

\subsection{Implementation Details}
Our WSDL approach consists of two subnetworks: multi-level attention extraction network (MAEN) and discriminative localization network (DLN). Both of them are all based on 16-layer VGGNet \cite{simonyan2014very}, which is widely used in image classification task. It can be replaced with the other CNNs. 
MAEN extracts the attention information of images to provide bounding boxes needed by DLN. For VGGNet in MAEN, we remove the layers after ``Conv$5\_3$'' and add a convolutional layer of size $3 \times 3$, stride $1$, pad $1$ with $1024$ neurons, which is followed by a GAP layer and a softmax layer \cite{zhou2016cvpr}. We adopt the object-level attention of Xiao et al. \cite{xiao2015application} to select relevant image patches for data extension. And then we utilize the extended data to fine-tune MAEN for learning discriminative features. The number of neurons in softmax layer is set as the number of subcategories in the dataset. DLN shares the weights of layers before ``Conv$5\_3$'' with MAEN for better discriminative localization as well as classification performance.

At training phase, for MAEN, we initialize the weights with the network pre-trained on the ImageNet 1K dataset \cite{imagenet_cvpr09}, and then use SGD with a minibatch size of 20. We use a weight decay of 0.0005 with a momentum of 0.9 and set the initial learning rate as 0.001. The learning rate is divided by 10 every 5K iterations. We terminate the training phase at 35K iterations on CUB-200-2011 dataset and 55K iterations on Cars-196 dataset because of different convergence rate. 
The discriminative localization network designed in our WSDL approach consists of one RPN and $n$ localization networks.
In the training phase, each localization network is trained one by one with RPN.
We first initialize the weights of the convolutional layers in RPN with the MAEN, and then train the RPN and 3 localization networks. When training the localization networks, the weights of RPN are fixed, and only the weights of the localization network are fine-tuned.
For the training of RPN and localization network, we start SGD with a minibatch size of 128, use a weight decay of 0.0005 with a momentum of 0.9 and set the initial learning rate to 0.001. We divide the learning
rate by 10 at 40K iterations on CUB-200-2011 dataset and 50K iterations on Cars-196 dataset, and terminate training at 90K iterations on CUB-200-2011 dataset and 120K iterations on Cars-196 dataset.

\subsection{Comparisons with State-of-the-art Methods}
This subsection presents the experimental results and analyses of our WSDL approach as well as the compared state-of-the-art methods on the widely-used CUB-200-2011 \cite{wah2011caltech} and Cars-196 \cite{krause20133d} datasets. We verify the effectiveness of our WSDL approach in the following aspects: (1) Accuracy of classification. (2) Efficiency of classification. The experimental results show that our WSDL achieves better performance than state-of-the-art methods in both accuracy and efficiency of classification.

\subsubsection{Accuracy of Classification}
Tables \ref{cub} and \ref{car} show the comparisons with state-of-the-art methods on CUB-200-2011 and Cars-196 datasets in the aspect of classification accuracy. Object, part annotations and CNN features used in these methods are listed for fair comparison. CNN model shown in the column of ``CNN Features'' indicates which CNN model is adopted to extract features. If the method adopt handcrafted feature like SIFT, the column of ``CNN Features'' is none. We present detailed analyses of our WSDL approach as well as compared methods on CUB-200-2011.

Early methods choose SIFT \cite{lowe2004distinctive} as basic feature and even use both object and part annotations, such as POOF \cite{berg2013poof} and HPM \cite{xie2013hierarchical}, but their classification results are limited and much lower than our WSDL approach. 
Our WSDL approach achieves the highest classification accuracy among all the state-of-the-art methods under the same weakly supervised setting, which indicates that neither object nor part annotations are used both in training and testing phases.
Our WSDL achieves the improvement by 1.02\% than the best state-of-the-art result of TSC \cite{he2017spatial} (85.71\% vs. 84.69\%), which jointly considers two spatial constraints in part selection. Despite achieving better classification accuracy, our WSDL approach is over two order of magnitude faster (i.e. 27 $\times$ faster) than TSC, as shown in Table \ref{cubtime}. The efficiency analyses will be described latter.
And our WSDL approach achieves better classification accuracy than the method of Bilinear-CNN \cite{lin2015bilinear}, which combines two different CNNs: VGGNet \cite{simonyan2014very} and VGG-M \cite{chatfield2014return}. Its classification accuracy is 84.10\%, lower than our approach by 1.61\%. 

Even compared with state-of-the-art methods using object annotations in both training and testing phases, such as Coarse-to-Fine \cite{yao2016coarse}, PG Alignment \cite{krause2015fine} and VGG-BGLm \cite{zhou2016fine}, our WSDL approach achieves improvement by at least 2.81\%. 
Moreover, our WSDL approach outperforms state-of-the-art methods using both object and part annotations, such as SPDA-CNN \cite{zhang2016spda}. 
Neither object nor part annotations are used in our WSDL approach, which marches toward practical application. Besides, the application of multi-level attention in our WSDL approach boosts the localization of discriminative regions and further improves the fine-grained image classification accuracy. 

The experimental results of comparisons with state-of-the-art methods on Cars-196 dataset in the aspect of classification accuracy are shown in Table \ref{car}. The trends are similar with CUB-200-2011 dataset, where our WSDL approach achieves the best classification accuracy among all the state-of-the-art methods under the same weakly supervised setting, which brings 1.00\% improvement than the best classification results from compared methods. Our WSDL approach outperforms those methods using object annotations, such as DPL-CNN \cite{wang2016weakly,zhou2016fine}, and is only beaten by PG Alignment \cite{krause2015fine} and BoT \cite{wang2016mining} no more than 0.30\%.

\begin{table*}[!ht]
  \centering
  \begin{tabular} {|p{3.5cm}<{\centering}|c|c|c|c|c|c|c|}
    \hline
    \multirow {2}{*}{Methods} & \multicolumn{2}{c|}{Train Annotation} & \multicolumn{2}{c|}{Test Annotation} & \multirow {2}{*}{Accuracy (\%)} & \multirow {2}{*}{CNN Features}\\
    \cline{2-5}
    &Object & Parts & Object & Parts & &\\
    \hline
    \textbf{Our WSDL Approach} & & & & & {\textbf{85.71}} & VGGNet \\
    \hline
    TSC \cite{he2017spatial}& & & & & 84.69 & VGGNet \\
    FOAF \cite{zhang2016fused} & & & & & 84.63 & VGGNet \\
    PD \cite{zhang2016picking}& & & & & 84.54 & VGGNet \\
    STN \cite{jaderberg2015spatial}& & & & & 84.10 & GoogleNet \\
    Bilinear-CNN \cite{lin2015bilinear}&  & &  & & 84.10 & VGGNet\&VGG-M \\
    PD (FC-CNN) \cite{zhang2016picking} & & & & & 82.60 & VGGNet \\
    Multi-grained \cite{wang2015multiple} & & & & & 81.70 & VGGNet \\
    NAC \cite{simon2015neural} & & & & & 81.01 & VGGNet \\
    PIR \cite{zhang2016weakly}& & & & & 79.34 & VGGNet \\
    TL Atten \cite{xiao2015application} & & & & & 77.90 & VGGNet \\
    MIL \cite{xu2017friend} & & & & & 77.40 & VGGNet \\
    VGG-BGLm \cite{zhou2016fine} & & & & & 75.90 & VGGNet \\
    InterActive \cite{xie2016interactive} & & & & & 75.62 & VGGNet \\
    \hline
    Coarse-to-Fine \cite{yao2016coarse}& $\surd$ & & $\surd$ & & 82.90 & VGGNet \\
    PG Alignment \cite{krause2015fine} & $\surd$ & & $\surd$ & & 82.80 & VGGNet \\
    Coarse-to-Fine \cite{yao2016coarse}& $\surd$ & & & & 82.50 & VGGNet \\
    VGG-BGLm \cite{zhou2016fine} & $\surd$ & & $\surd$ & & 80.40 & VGGNet \\
    Triplet-A (64) \cite{cui2015fine} & $\surd$ & & $\surd$ & & 80.70 & GoogleNet \\
    Triplet-M (64) \cite{cui2015fine} & $\surd$ & & $\surd$ & & 79.30 & GoogleNet \\
    \hline
    Webly-supervised \cite{xu2016webly} & $\surd$ & $\surd$ &  &  & 78.60 & AlexNet \\
    PN-CNN \cite{branson2014bird} & $\surd$ & $\surd$ &  &  & 75.70 & AlexNet \\
    Part-based R-CNN \cite{zhang2014part} & $\surd$ & $\surd$ &  & & 73.50 & AlexNet \\
    SPDA-CNN \cite{zhang2016spda} & $\surd$ & $\surd$ & $\surd$ &  & 85.14 & VGGNet \\
    Deep LAC \cite{lin2015deep} & $\surd$ & $\surd$ & $\surd$ &  & 84.10 & AlexNet \\
    SPDA-CNN \cite{zhang2016spda} & $\surd$ & $\surd$ & $\surd$ &  & 81.01 & AlexNet \\
    Part-stacked CNN \cite{huang2016part}& $\surd$ & $\surd$ & $\surd$ &  & 76.20 & AlexNet \\
    PN-CNN \cite{branson2014bird} & $\surd$ & $\surd$ & $\surd$ & $\surd$ & 85.40 & AlexNet  \\
    Part-based R-CNN \cite{zhang2014part} & $\surd$ & $\surd$ & $\surd$ & $\surd$ & 76.37 & AlexNet \\
    POOF \cite{berg2013poof} & $\surd$ & $\surd$ & $\surd$ & $\surd$ & 73.30 &  \\
    HPM \cite{xie2013hierarchical} & $\surd$ & $\surd$ & $\surd$ & $\surd$ & 66.35 &  \\
    \hline
  \end{tabular}
  \caption{Comparisons with state-of-the-art methods on CUB-200-2011 dataset in the aspect of classification accuracy to show the effectiveness of our WSDL approach.}\
  \label{cub}
\end{table*}

\begin{table*}[!ht]
  \centering
  \begin{tabular} {|p{3.5cm}<{\centering}|c|c|c|c|c|c|}
    \hline
    \multirow {2}{*}{Methods} & \multicolumn{2}{c|}{Train Annotation} & \multicolumn{2}{c|}{Test Annotation} & \multirow {2}{*}{Accuracy (\%)} & \multirow {2}{*}{CNN Features}\\
    \cline{2-5}
    &Object & Parts & Object & Parts & &\\
    \hline
    \textbf{Our WSDL Approach} & & & & & {\textbf{92.30}} & VGGNet \\
    \hline
    Bilinear-CNN \cite{lin2015bilinear} & & & & & 91.30 & VGGNet\&VGG-M \\
    TL Atten \cite{xiao2015application}& & & & & 88.63 & VGGNet\\
    DVAN \cite{zhao2016diversified} & & & & & 87.10 & VGGNet \\
    FT-HAR-CNN \cite{xie2015hyper} & & & & & 86.30 & AlexNet \\
    HAR-CNN \cite{xie2015hyper} & & & & & 80.80 & AlexNet \\
    \hline
    PG Alignment \cite{krause2015fine}& $\surd$ & & & & 92.60 & VGGNet \\
    ELLF \cite{krause2014learning} & $\surd$ & & & & 73.90 & CNN \\
    R-CNN \cite{girshick2014rich} & $\surd$ & & & & 57.40 & AlexNet \\
    PG Alignment \cite{krause2015fine} & $\surd$ & & $\surd$ & & 92.80 & VGGNet \\
    BoT(CNN With Geo) \cite{wang2016mining}& $\surd$ & & $\surd$ & & 92.50 & VGGNet \\
    DPL-CNN \cite{wang2016weakly}& $\surd$ & & $\surd$ & & 92.30 & VGGNet \\
    VGG-BGLm \cite{zhou2016fine} & $\surd$ & & $\surd$ & & 90.50 & VGGNet \\
    BoT(HOG With Geo) \cite{wang2016mining}& $\surd$ & & $\surd$ & & 85.70 & VGGNet \\
    LLC \cite{wang2010locality} & $\surd$ & & $\surd$ & & 69.50 &  \\
    BB-3D-G \cite{krause20133d} & $\surd$ & & $\surd$ & & 67.60 &  \\
    \hline
  \end{tabular}
  \caption{Comparisons with state-of-the-art methods on Cars-196 dataset in the aspect of classification accuracy to show the effectiveness of our WSDL approach.}
  \label{car}
\end{table*}

\begin{table*}[!ht]
  \centering
  \begin{tabular} {|p{5.2cm}<{\centering}|p{4cm}<{\centering}|p{2.7cm}<{\centering}|}
    \hline
    Methods & Average Classification Speed (fps)  & CNN Models\\
    \hline
    \textbf{Our WSDL Approach} & \textbf{9.09} &  VGGNet \\
    Bilinear-CNN \cite{lin2015bilinear}& 4.52 &  VGGNet\&VGG-M \\
    TSC \cite{he2017spatial} & 0.34 &  VGGNet \\
    TL Atten \cite{xiao2015application} & 0.25 &VGGNet \\
    NAC \cite{simon2015neural} & 0.10  &VGGNet \\
    \hline
    \textbf{Our WSDL Approach}  & \textbf{16.13} & AlexNet \\
    Part-stacked CNN \cite{huang2016part} & 14.30 & AlexNet\\
    \hline
  \end{tabular}
  \caption{Comparisons with state-of-the-art methods in the aspect of classification efficiency. CUB-200-2011 dataset is adopted as the evaluation dataset, and average classification speed is evaluated by the frames recognized per second, denoted as fps. The results are all obtained on the computer with one GPU of NVIDIA TITAN X @1417MHZ and one CPU of Intel Core i7-6900K @3.2GHZ.}\
  \label{cubtime}
\end{table*}
\subsubsection{Efficiency of Classification}
Experimental results of comparisons with state-of-the-art methods in the aspect of classification efficiency is presented in Table \ref{cubtime}. 
Average classification speed is evaluated by the frames recognized per second, denoted as fps.
Since it has little relation with datasets, CUB-200-2011 dataset is adopted as the evaluation dataset. We get the average classification speed on the computer with one GPU of NVIDIA TITAN X @1417MHZ and one CPU of Intel Core i7-6900K @3.2GHZ. 
Compared with state-of-the-art methods, our WSDL approach achieves the best performance on not only the classification accuracy but also the efficiency. 
We split state-of-the-art methods into 2 groups by the basic CNN models used in their methods: VGGNet \cite{simonyan2014very} and AlexNet \cite{krizhevsky2012imagenet}. 
Apart from hardware environment, average classification speed also depends on implementation of the method. Different implementations achieve different average classification speeds.
For fair comparison, we directly run the source codes provided by authors of compared methods under the same experimental setting, except Part-stacked CNN \cite{huang2016part}. Its average classification speed is reported as 20 fps in the original paper. It reports that a single CaffeNet \cite{jia2014caffe} runs at 50 fps under the experimental setting (NVIDIA Tesla K80). In our experiments, a single CaffeNet runs at 35.75 fps, so we calculate the speed of Part-stacked CNN in the same experimental setting with ours as 20$\times$35.75$\div$50=14.30 fps.
Compared with state-of-the-art methods in the first group, our WSDL approach is 2 $\times$ faster than Bilinear-CNN (9.09 fps vs. 4.52 fps). Besides, the classification accuracy of our WSDL approach is also 1.61\% higher than Bilinear-CNN on CUB-200-2011 dataset.  
Even more, our WSDL approach is over two orders of magnitude faster than methods based on labor-consuming region proposal methods, such as TSC \cite{he2017spatial}, TL Atten \cite{xiao2015application} and NAC \cite{simon2015neural}.
When utilizing AlexNet as the basic CNN, our WSDL approach is still faster than Part-stacked CNN \cite{huang2016part}, which also utilizes AlexNet.
It is noted that neither object nor part annotations are used in our approach, while all are used in Part-stacked CNN. 
Our WSDL approach avoids the time-consuming classification process by the design of discriminative localization network (DLN) with one region proposal network and multiple localization networks, and achieves the best classification accuracy by the mutual promotion between localization and classification. This leads the fine-grained image classification to practical application. 

\subsection{Effectivenesses of Components in Our WSDL Approach}

Detailed experiments are performed to show the effectiveness of each component in our WSDL approach in the following two aspects:
\subsubsection{Effectiveness of multi-level attention in the aspect of classification accuracy}
In our WSDL approach, multi-level attention is applied. Different level attentions focus on different characteristics of the image, which are complementary and boost the classification accuracy. In the experiments, we extract the attention maps from the convolutional layers of ``Conv4\_3'', ``Conv5\_3'' and ``Conv\_cam'' in our MAEN, and evaluate their effectivenesses.
From Table \ref{multilevel}, we can observe that the combination of different level attentions boosts the classification accuracy, which verifies the complementarity among them. The attention from ``Conv4\_3'' plays a minor role in promoting the classification accuracy. Besides, the time consumption of the application of three-level attention is high. Therefore, in our experiments, we only adopt two-level attention from ``Conv5\_3'' and ``Conv\_cam'' to achieve the best trade-off between classification accuracy and efficiency, as shown in Tables \ref{cub} to \ref{cubtime}. 

\subsubsection{Effectiveness of discriminative localization network in the aspect of classification efficiency}
Since we apply multi-level attention, there are 2 choices: (1) Train $n$ discriminative localization networks, each of which consists of one RPN and one Fast R-CNN, denoted as ``two-level (respectively)'' and ``three-level (respectively)'' in Table \ref{dlntime}, which causes the linear growth of time consumption.
(2) In our WSDL approach, we design an $n$-pathway discriminative localization network with one RPN and $n$ localization networks, and all of them share the same region proposals generated by RPN, which avoids the linear growth of time consumption, denoted as ``two-level (dln)'' and ``three-level (dln)'' in Table \ref{dlntime}. From Table \ref{dlntime}, we can observe that our designed architecture of DLN reduces the time consumption.

\begin{figure*}[!ht]
    \begin{center}\includegraphics[width=1\linewidth]{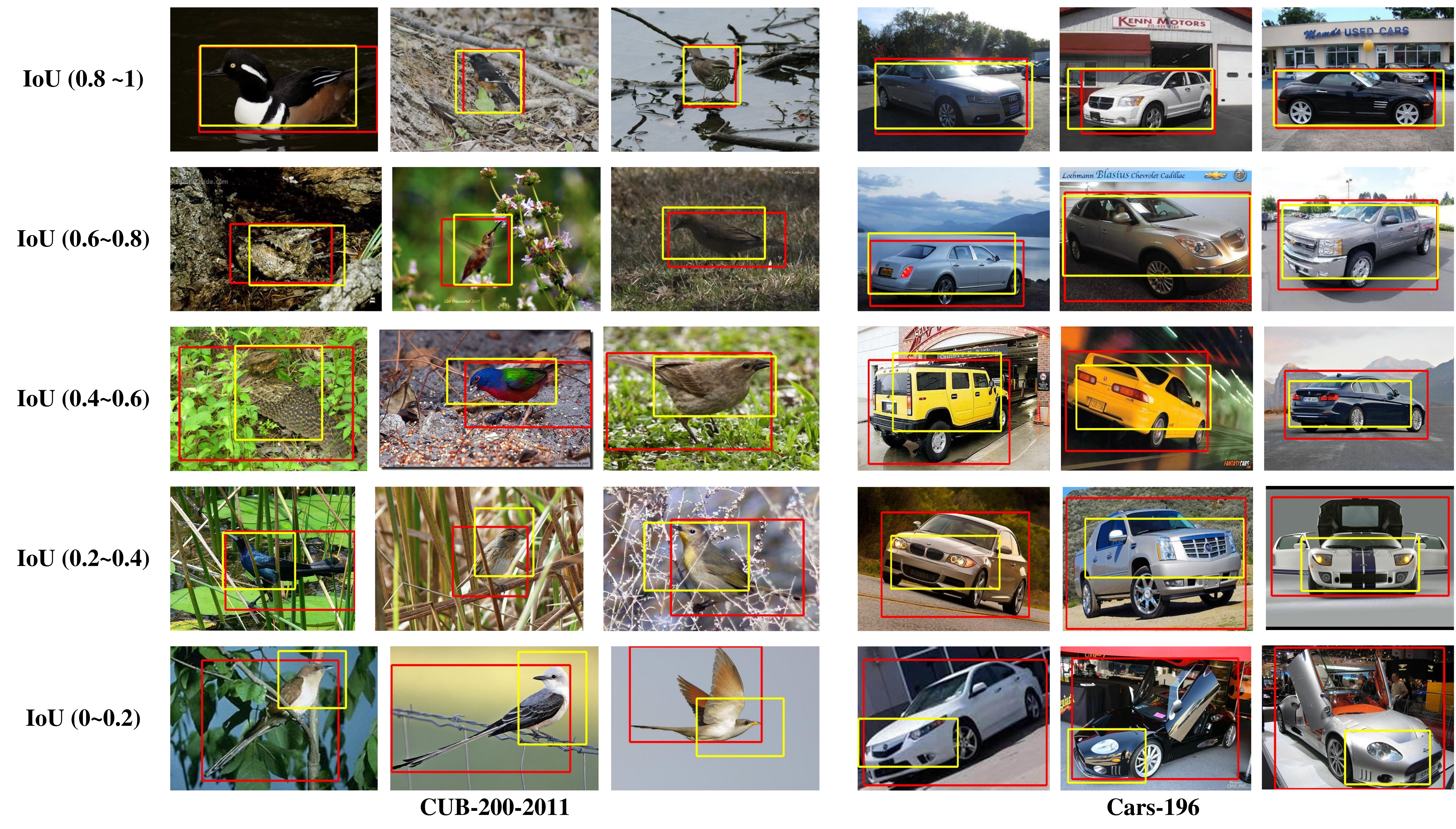}
    \caption{Samples of predicted bounding boxes of discriminative regions (yellow rectangles) based on the attention information from ``Conv\_cam'' and ground truth bounding boxes of objects (red rectangles) at different ranges of IoU on CUB-200-2011 and Cars-196 datasets.}
    \label{boundingbox}
    \end{center}
\end{figure*}
\begin{figure}[!ht]
    \begin{center}\includegraphics[width=0.9\linewidth]{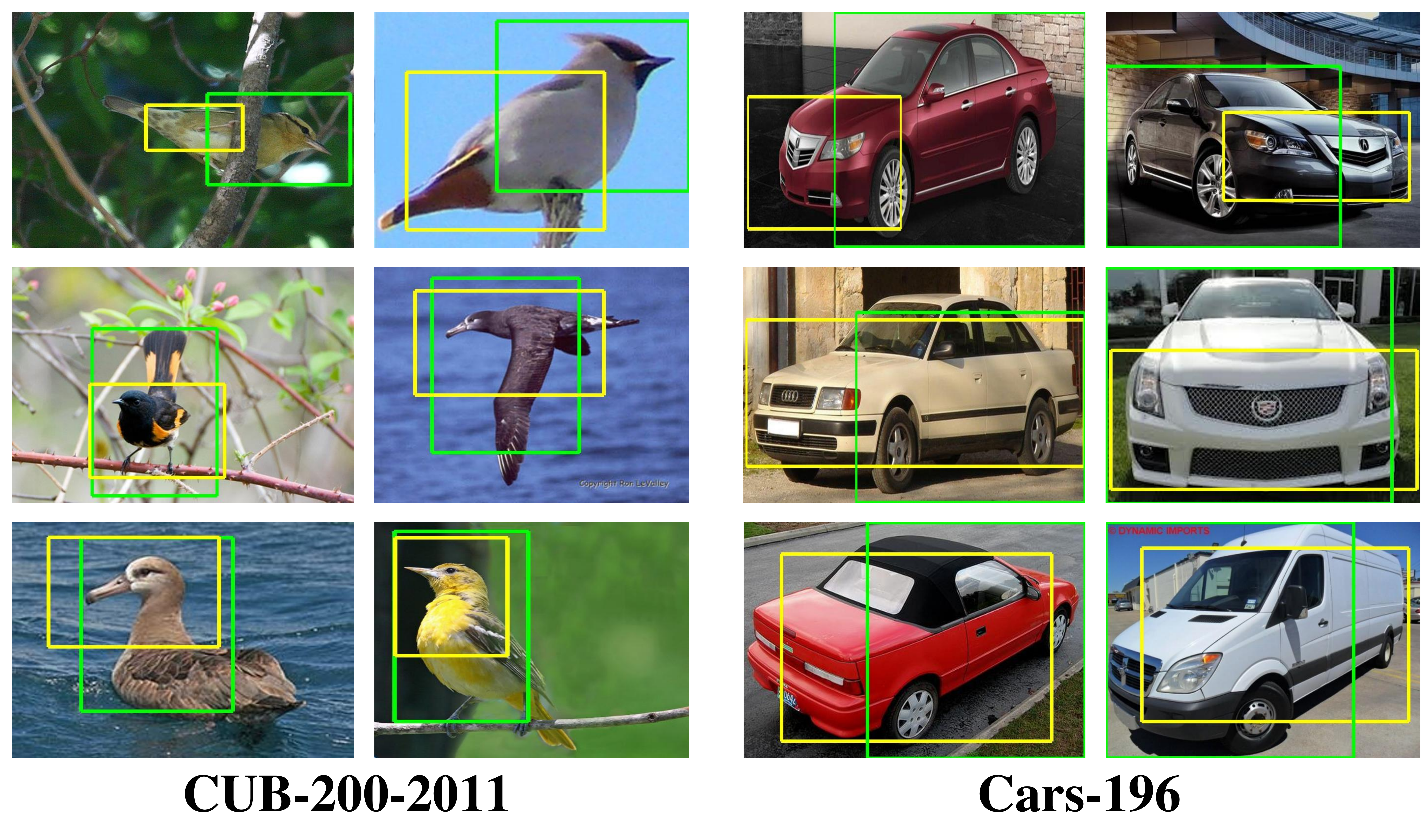}
    \caption{Samples of the bounding boxes of discriminative regions based on two level attentions from ``Conv\_cam'' (yellow rectangles) and ``Conv5\_3'' (green rectangles) on CUB-200-2011 and Cars-196 datasets.}
    \label{boundingbox-2level}
    \end{center}
\end{figure}
\subsection{Comparisons with Baselines}
Our WSDL approach is based on Faster R-CNN \cite{ren2015faster}, MAEN, and VGGNet \cite{simonyan2014very}. To verify the effectiveness of our WSDL approach, we present the results of comparisons with Faster R-CNN, MEAN and VGGNet on CUB-200-2011 dataset in Table \ref{baselines}. ``VGGNet'' denotes the result of directly using fine-tuned VGGNet, ``MEAN'' denotes the result of directly using MEAN, and ``Faster R-CNN (gt)'' denotes the result of directly adopting Faster R-CNN with ground truth bounding box of the object to guide training phase. Our WSDL approach achieves the best performance even without using object or part annotations. We adopt VGGNet as the basic model in our approach, but its classification accuracy is only 70.42\%, which is much lower than ours.
It shows that the discriminative localization has promoting effect to classification. 
With discriminative localization, we find the most important regions of images for classification, which contain the key variance from other subcategories. Compared with ``Faster R-CNN (gt)'', our approach also achieves better performance. It is an interesting phenomenon that worth thinking about. From the last row in Fig. \ref{boundingbox}, we observe that not all the areas in the ground truth bounding boxes are helpful for classification.
Some ground truth bounding boxes contain large area of background noise that has less useful information and even leads to misclassification. So discriminative localization is significantly helpful for achieving better classification performance.  

\subsection{Effectiveness of Discriminative Localization}
\begin{table}[!ht]
  \centering
   \begin{tabular} {|c|c|c|}
    \hline
    \multirow {2}{*}{Convolutional layers} & \multicolumn{2}{c|}{Accuracy (\%)} \\
    \cline{2-3}
        &CUB-200-2011&Cars-196\\
    \hline
    Conv\_cam & 83.45 & 89.59 \\
    Conv5\_3 & 81.15 & 84.31 \\
    Conv4\_3 & 77.84 & 78.01 \\
    \hline
    Conv\_cam + Conv5\_3 & 84.43 & 90.29 \\
    Conv\_cam + Conv4\_3 & 84.36 & 90.10 \\
    Conv4\_3 + Conv5\_3 & 81.41 & 84.68 \\
    \hline
    Conv\_cam + Conv5\_3 + Conv4\_3 & 84.59 & 90.30 \\
    \hline
  \end{tabular}
  \caption{Effectiveness of multi-level attention in our WSDL approach on CUB-200-2011 and Cars-196 datasets in the aspect of classification accuracy. 
  }
  \label{multilevel}
\end{table}
\begin{table}[!ht]
  \centering
   \begin{tabular} {|p{3.1cm}<{\centering}|p{4cm}<{\centering}|}
    \hline
    Methods & Average Classification Speed (fps) \\
    \hline
    one-level & 10.07 \\
    two-level (respectively) & 5.04 \\
    two-level (with dln) & 9.09 \\
    three-level (respectively)& 3.36  \\
    three-level (with dln)& 7.69  \\
    \hline
  \end{tabular}
  \caption{Effectiveness of discriminative localization network in our WSDL approach on CUB-200-2011 dataset in the aspect of classification efficiency. 
  }
  \label{dlntime}
\end{table}
\begin{table}[!t]
  \centering
  \begin{tabular} {|p{3.1cm}<{\centering}|p{4cm}<{\centering}|}
    \hline
    Methods & Accuracy (\%) \\
    \hline
    \textbf{Our WSDL Approach }  & \textbf{85.71} \\
    \hline
    Faster R-CNN (gt) & 82.46 \\
    MEAN & 77.50 \\
    VGGNet & 70.42 \\
    \hline
  \end{tabular}
   \caption{Comparison with baselines on CUB-200-2011 dataset. }
  \label{baselines}
\end{table}
\begin{table}[!t]
  \centering
  \begin{tabular} {|p{3cm}<{\centering}|p{1.8cm}<{\centering}|p{1.8cm}<{\centering}|}
    \hline
    \multirow {2}{*}{Methods} & \multicolumn{2}{c|}{Localization Accuracy (\%)} \\
    \cline{2-3}
        &CUB-200-2011&Cars-196\\
    \hline
    \textbf{Our WSDL Approach }  & \textbf{46.05} & \textbf{56.60} \\
    MEAN & 44.93 & 55.79\\
    \hline
  \end{tabular}
   \caption{Localization results on CUB-200-2011 and Cars-196 datasets. }
  \label{localization}
\end{table}
Our WSDL approach focuses on improving the localization and classification performance simultaneously. Since the discriminative regions are generally located at the region of the object in the image, we adopt the IoU overlap between the discriminative region and ground truth bounding box of the object to evaluate the correctness of localization. We consider a bounding box of discriminative region to be correctly predicted if its IoU with ground truth bounding box of the object is larger than 0.5. We show the results obtained from ``Conv\_cam'' on CUB-200-2011 and Cars-196 datasets in Table \ref{localization}, and our WSDL approach achieves the accuracy of 46.05\% and 56.60\%. Considering that neither object nor part annotations are used, it is a promising result. Compared with ``MAEN'' which means directly using the attention map from ``Conv\_cam'' to generate bounding box, our WSDL approach achieves improvements by 8.21\%, which verifies its effectiveness.
\begin{table*}[!ht]
  \centering
  \begin{tabular} {|p{1.47cm}<{\centering}|p{1.47cm}<{\centering}|p{1.47cm}<{\centering}|p{1.47cm}<{\centering}|p{1.47cm}<{\centering}|p{1.47cm}<{\centering}|p{1.47cm}<{\centering}|p{1.47cm}<{\centering}|p{1.47cm}<{\centering}|}
    \hline
    Parts & back & beak & belly & breast & crown & forehead & left eye & left leg\\
    \hline
    PCL (\%) & 96.33 & 96.49 & 94.00 & 95.29 & 97.38 & 97.07 & 97.49 & 89.92 \\
    \hline
    Parts & left wing & nape & right eye & right leg & right wing & tail & throat & \textbf{average}\\
    \hline
    PCL (\%)& 92.60 & 96.60 & 96.79 & 91.85 & 97.00 & 85.03 & 96.38 & \textbf{94.68}\\
    \hline
  \end{tabular}
  \caption{PCL for each part of the object in the CUB-200-2011 dataset.}
  \label{parts}
\end{table*}
\begin{figure*}[!t]
    \begin{center}\includegraphics[width=1\linewidth]{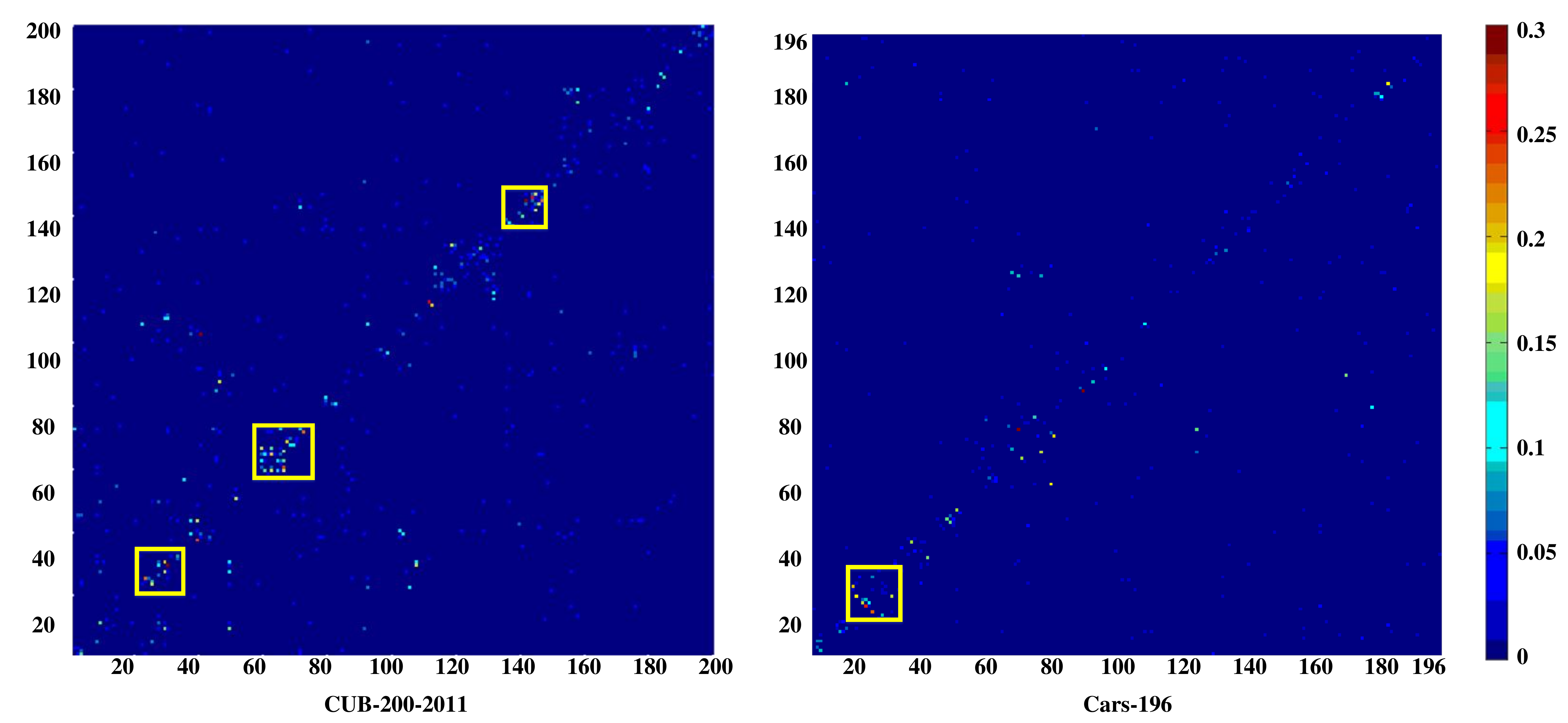}
    \caption{Classification confusion matrices on CUB-200-2011 and Cars-196 datasets. The yellow rectangles show the sets of subcategories with the higher probability of misclassification. }
    \label{confusion}
    \end{center}
\end{figure*}
\begin{figure*}[!t]
    \begin{center}\includegraphics[width=1\linewidth]{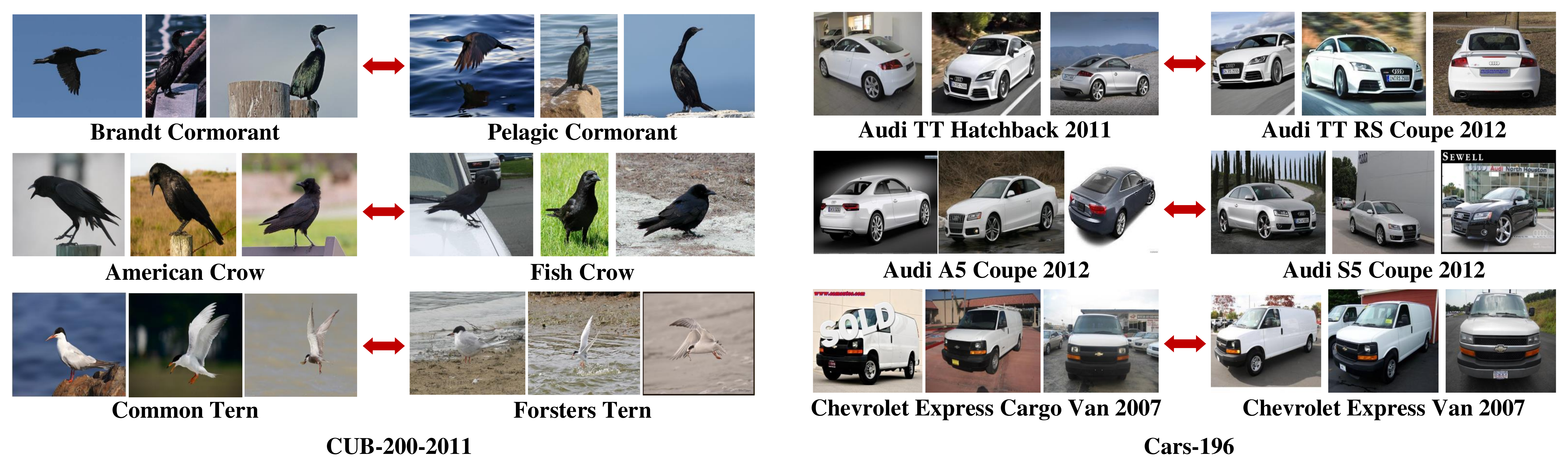}
    \caption{Examples of the most confused subcategory pairs in CUB-200-2011 and Cars-196 datasets. One subcategory is mostly confidently classified as the other in the same row in the testing phase.}
    \label{confusionexamples}
    \end{center}
\end{figure*}
We also show some samples of predicted bounding boxes of discriminative regions and ground truth bounding boxes of objects at different ranges of IoU (e.g. 0$\sim$0.2, 0.2$\sim$0.4, 0.4$\sim$0.6, 0.6$\sim$0.8, 0.8$\sim$1) on CUB-200-2011 and Cars-196 datasets, as shown in Fig. \ref{boundingbox}. We have some predicted bounding boxes whose IoUs with ground truth bounding boxes of objects are lower than 0.5. But these predicted bounding boxes contain discriminative regions of the objects, such as heads or bodies. 
It verifies the effectiveness of our WSDL approach in localizing discriminative regions of object for achieving better classification performance. 
Fig. \ref{boundingbox-2level} shows the bounding boxes of discriminative regions based on two level attentions from ``Conv\_cam'' and ``Conv5\_3''. We can observe that different attentions focus on different regions, and provide complementary information to boost the classification accuracy.
To further verify the effectiveness of discriminative localization in our WSDL approach, quantitative results are given in terms of the  Percentage of Correctly Localization (PCL) in Table \ref{parts}, which estimates whether the predicted bounding box contains the parts of object or not. CUB-200-2011 dataset provides 15 part locations, which denote the pixel locations of centers of parts. We consider our predicted bounding box contains a part if the part location lies in the area of the predicted bounding box. Table \ref{parts} shows that about average 94.68\% of the parts located in our predicted bounding boxes. It shows that our discriminative localization can detect the distinguishing information of objects to promote classification performance.

\subsection{Analysis of Misclassification} 
Fig. \ref{confusion} shows the classification confusion matrices on CUB-200-2011 and Cars-196 datasets, where coordinate axes denote subcategories and different colors denote different probabilities of misclassification. The yellow rectangles show the sets of subcategories with the higher probability of misclassification. We can observe that these sets of subcategories locate near the diagonal of the confusion matrices, which means that these misclassification subcategories generally belong to the same genus or car brand with small variance. The small variance is not easy to measure from the image, which leads the high challenge of fine-grained image classification. 
Fig. \ref{confusionexamples} shows some examples of the most probably confused subcategory pairs. One subcategory is most confidently classified as the other in the same row. The subcategories in the same row look almost the same, and belong to the same genus. For example, ``Common Tern'' and ``Forsters Tern'' look the same in the appearance, as shown in the left third row of Fig. \ref{confusionexamples}, because both of them have the same attributes of white wings and black forehead, and belong to the genus of ``Tern''. It is even extremely difficult for human beings to distinguish between them. Similarly, it is hard to distinguish between ``Audi TT Hatchback 2011'' and ``Audi TT RS Coupe 2012''.

\section{Conclusion}
In this paper, the weakly supervised discriminative localization approach (WSDL) has been proposed for fast fine-grained image classification. 
We first apply multi-level attention to guide the discriminative localization learning to localize multiple discriminative regions simultaneously for each image, which only uses image-level subcategory label to avoid using labor-consuming annotations. 
Then we design an $n$-pathway end-to-end discriminative localization network to simultaneously localize discriminative regions and encode discriminative features, which not only achieves a notable classification performance but also improves classification speed. Their combination simultaneously improves classification speed and eliminates dependence on object and part annotations.  
Comprehensive experimental results show our WSDL approach is more effective and efficient compared with state-of-the-art methods on 2 widely-used datasets.  

The future works lie in three aspects: 
First, we will make our WSDL approach better by learning better discriminative localization via exploiting the effectiveness of fully convolutional networks. Second, we will make our WSDL approach faster by designing a more efficient network with less operations for a forward pass. Third, we will make our WSDL approach more generalized by training one model to support the classification of different datasets.




\ifCLASSOPTIONcaptionsoff
  \newpage
\fi



%
\bibliographystyle{plain}
\bibliography{sigproc}
\end{document}